\documentclass[sigconf,natbib=true]{acmart}

\usepackage{microtype}
\usepackage{booktabs}
\usepackage{multirow}
\usepackage{multicol}
\usepackage{makecell}
\usepackage{enumitem}
\usepackage{graphicx}
\usepackage{subfigure}
\usepackage{caption}
\usepackage{amsmath}
\usepackage{amsfonts}
\usepackage{color}

\AtBeginDocument{%
  \providecommand\BibTeX{{%
    \normalfont B\kern-0.5em{\scshape i\kern-0.25em b}\kern-0.8em\TeX}}}

\setcopyright{acmcopyright}
\copyrightyear{2022}
\acmYear{2022}

\acmConference[XXXX]{ACM Conference}{August 2022}{XXX}%
\acmBooktitle{}
\acmPrice{15.00}
\acmISBN{XXX-X-XXXX-XXXX-X/22/08}

\begin{document}

\title[Follow Me: Conversation Planning for Target-driven Recommendation Dialogue Systems]{Follow Me: Conversation Planning for Target-driven Recommendation Dialogue Systems}


%
\author{Jian Wang}
\affiliation{
   \institution{The Hong Kong Polytechnic University}
   \country{}
}
\email{csjiwang@comp.polyu.edu.hk}

\author{Dongding Lin}
\affiliation{
   \institution{The Hong Kong Polytechnic University}
   \country{}
}
\email{csdlin@comp.polyu.edu.hk}

\author{Wenjie Li}
\affiliation{
   \institution{The Hong Kong Polytechnic University}
   \country{}
}
\email{cswjli@comp.polyu.edu.hk}


\renewcommand{\shortauthors}{Wang et al.}

\begin{abstract}
Recommendation dialogue systems aim to build social bonds with users and provide high-quality recommendations. This paper pushes forward towards a promising paradigm called target-driven recommendation dialogue systems, which is highly desired yet under-explored. We focus on how to naturally lead users to accept the designated targets gradually through conversations. To this end, we propose a Target-driven Conversation Planning (TCP) framework to plan a sequence of dialogue actions and topics, driving the system to transit between different conversation stages proactively. We then apply our TCP with planned content to guide dialogue generation. Experimental results show that our conversation planning significantly improves the performance of target-driven recommendation dialogue systems.
\end{abstract}

%
%


\maketitle

\section{Introduction}
In recent years, an important special type of task-oriented dialogue systems named recommendation dialogue systems \cite{chen-etal-2019-towards, kang-etal-2019-recommendation} has gained growing research interest, which is expected to encourage natural interactions with users so as to make better recommendations. It reveals that recommendation-oriented tasks are beneficial to deeply tap the application potential of dialogue systems \cite{jannach2021survey}.

\begin{figure}[th!]
\centering
\includegraphics[width=0.95\linewidth]{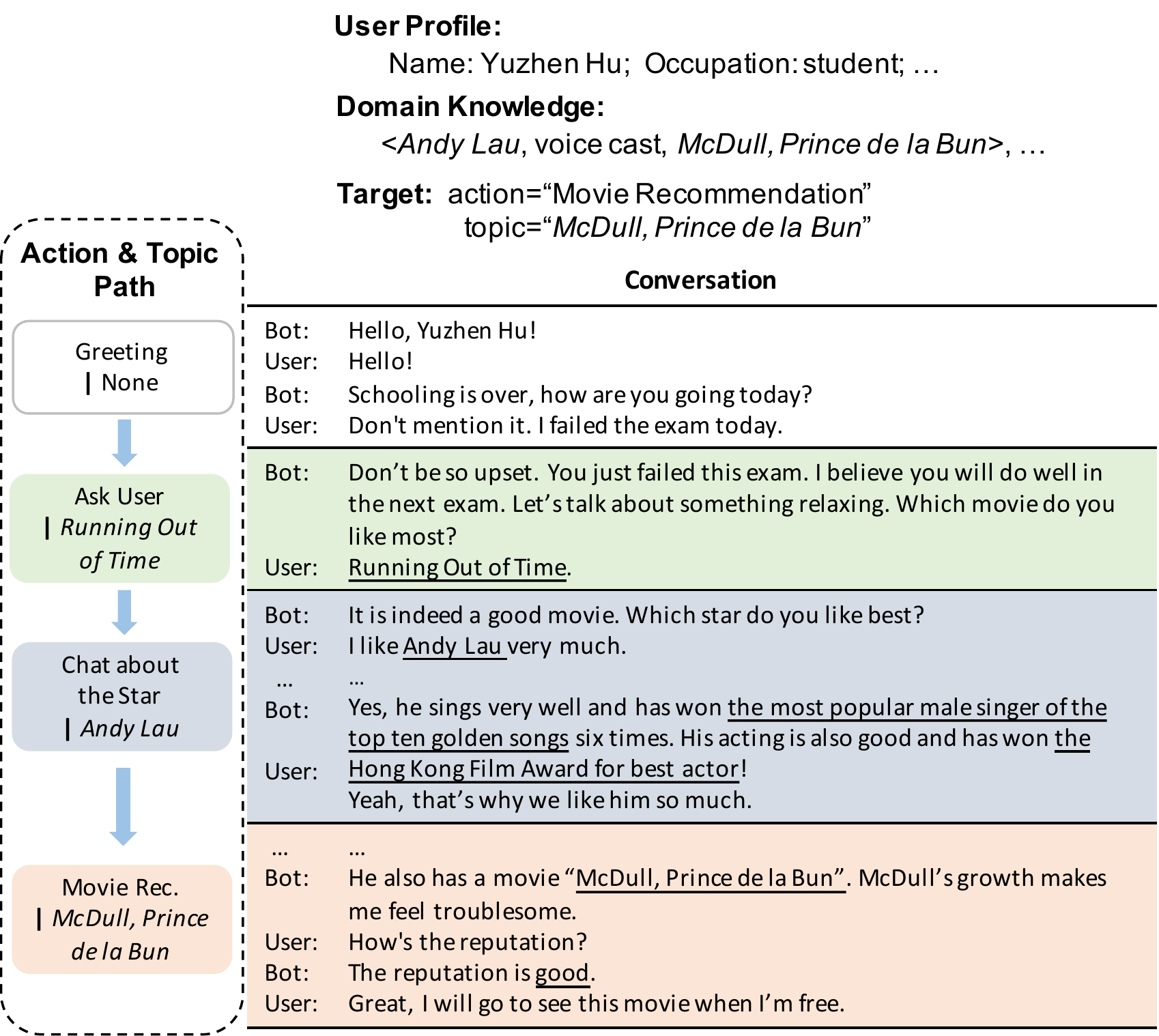}
\caption{An illustrative example from the re-purposed DuRecDial \cite{liu-etal-2020-towards-conversational} dataset. The whole conversation is grounded on the user profile, domain knowledge, and the target.}
\label{fig:example}
\end{figure}

It was the emergence of multiple datasets that helps push forward the research in this area, such as \textsc{GoRecDial} \cite{kang-etal-2019-recommendation}, TG-ReDial \cite{zhou-etal-2020-towards}, \textsc{INSPIRED} \cite{hayati-etal-2020-inspired}. As follow-up studies, \citet{ma-etal-2021-cr} proposed a tree-structured reasoning framework over knowledge graphs to guide both item recommendations and response generations. \citet{liang-etal-2021-learning} introduced a NTRD framework to combine the advantage of classic slot filling and neural language generation for item recommendations. However, most existing recommendation dialogue systems \cite{chen-etal-2019-towards, kang-etal-2019-recommendation,ma-etal-2021-cr,liang-etal-2021-learning} converse with users \textit{reactively}. They mainly respond to users' utterances in order to better understand the expressed preferences or requirements, and then provide recommendations accordingly. Such reactive recommendation dialogue systems have their limitation in reality since people may not have clear preferences for the unfamiliar new topics or items. 

We are desired to explore how to \textit{proactively} recommend target topics or items that possibly attract users through conversations in more sociable ways. Recently, the emergence of the DuRecDial \cite{liu-etal-2020-towards-conversational} dataset has shed light on this research direction. As shown in the example of Figure \ref{fig:example}, suppose there is a target movie named ``\textit{McDull, Prince de la Bun}'', the system (i.e., Bot) is required to proactively and naturally lead the whole conversation (e.g., ``\textit{greeting}'' $\rightarrow$ ``\textit{ask user}'' $\rightarrow$ ``\textit{chat about the star}'' $\rightarrow$ ``\textit{movie recommendation}'') so as to recommend the target movie when appropriate.
To accomplish the above process, the system needs to consider the user profile, the domain knowledge, and the target for generating system utterances.
Specifically, the user profile is important for the system to take initiative and warm up a conversation since it reveals the user's attributes and past preferences. The domain knowledge about domain-specific topics and associated attributes is also crucial to enable smooth topic transitions (e.g., ``\textit{Running Out of Time}'' $\rightarrow$ ``\textit{Andy Lau}'' $\rightarrow$ ``\textit{McDull, Prince de la Bun}'').

\begin{figure}[th!]
\centering
\subfigure[Multi-task learning paradigm]{
    \begin{minipage}[b]{0.4\linewidth}
	    \includegraphics[width=1\linewidth]{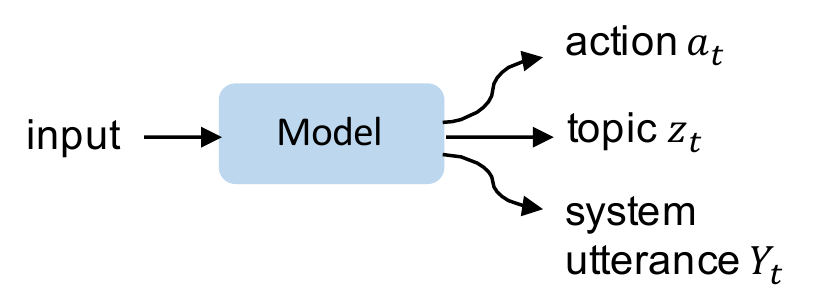}
    \end{minipage}
    \label{fig:comp.a}
}
\subfigure[Predict-then-generate paradigm]{
    \begin{minipage}[b]{0.55\linewidth}
	    \includegraphics[width=1\linewidth]{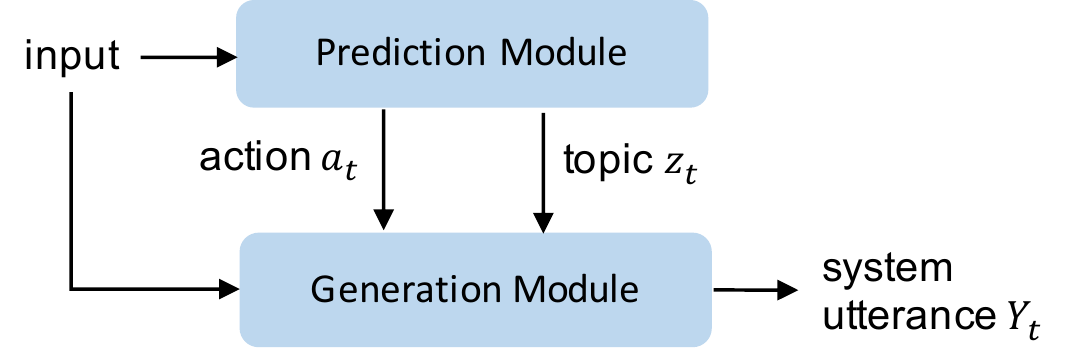}
    \end{minipage}
    \label{fig:comp.b}
}
\subfigure[Our target-driven planning enhanced generation framework]{
\includegraphics[width=0.75\linewidth]{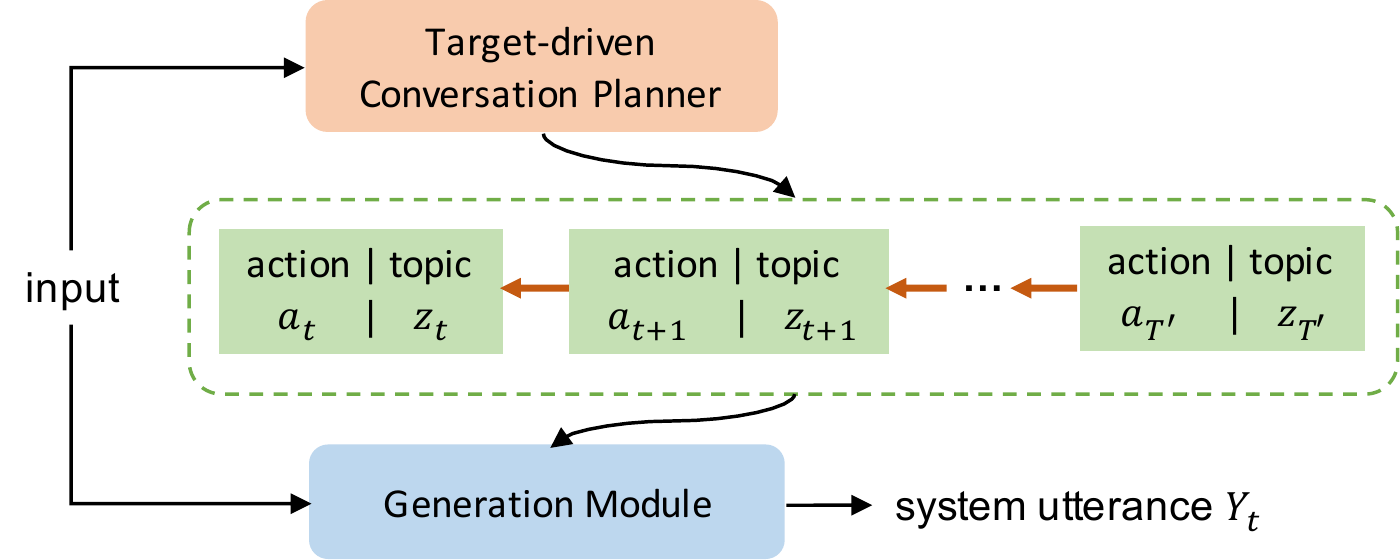}
\label{fig:comp.c}
}
\caption{Comparison of different paradigms.}
\label{fig:comparison}
\end{figure}

In this paper, we move forward to target-driven recommendation dialogue systems. Given a designated target topic (e.g., movie, music, food), a dialogue system is expected to proactively lead the conversation towards its target in order to make a successful recommendation. Compared to previous recommendation-oriented dialogues, our key research question is ``\textit{How to make reasonable plans to drive the conversation to reach the designated target step by step?}''. It is challenging because (1) the system should always maintain an engaging conversation to attract the user's attention and smoothly transit among relevant topics, and (2) the system is required to be able to arouse the user’s interest in the target topic to be recommended rather than discovering user preferences alone.

Although there are related works using the multi-task learning \cite{lin2021target} paradigm (Figure \ref{fig:comp.a}) and the predict-then-generate \cite{liu-etal-2020-towards-conversational,zhang-etal-2021-kers} paradigm (Figure \ref{fig:comp.b}), we propose a different framework named \textbf{T}arget-driven \textbf{C}onversation \textbf{P}lanning (\textbf{TCP}) to guide dialogue generation (Figure \ref{fig:comp.c}), which aims to plan a path consisting of dialogue topics and the ways how the system delivers these topics (i.e., dialogue actions). The key module is the target-driven conversation planner, which is based on the widely-used Transformer \cite{vaswani2017attention} network. We use the planned content to help extract necessary knowledge and explicitly guide the system to generate utterances.

The main contributions of this paper are summarized in two folds. (1) To the best of our knowledge, we are the first to push forward from the reactive recommendation dialogue paradigm towards the promising proactive paradigm by designating targets and formulating the target-driven recommendation dialogue task. (2) We propose a Target-driven Conversation Planning (TCP) framework to plan a path of dialogue actions and topics, which helps the system to lead the conversation and guide the utterance generation.

\section{Method}

\subsection{Problem Formulation}
Suppose we have a recommendation-oriented dialogue corpus $\mathcal{D}=\{(\mathcal{U}_{i},\mathcal{K}_{i},\mathcal{H}_{i},\mathcal{P}_{i})\}_{i=1}^{N}$, where $\mathcal{U}_{i}=\{u_{i,j}\}_{j=1}^{N_U}$ denotes a user profile with each entry $u_{i,j}$ in form of a $\langle\textit{key, value}\rangle$ pair, $\mathcal{K}_{i}=\{k_{i,j}\}_{j=1}^{N_K}$ denotes a set of domain knowledge facts relevant to $i$-th conversation with each element $k_{i,j}$ in form of a $\langle\textit{subject, relation, object}\rangle$ triple, $\mathcal{H}_{i}=\{(X_{i,t},Y_{i,t})\}_{t=1}^{T}$ denotes conversation content with a total number of $T$ turns, $\mathcal{P}_{i}=\{(a_{i,l}, z_{i,l})\}_{l=1}^{L}$ denotes a sequence of annotated plans and each plan specifies a dialogue action $a_{i,l}$ and a dialogue topic $z_{i,l}$. Here, the dialogue topics are mainly constructed upon the domain knowledge $\mathcal{K}_{i}$, each action/topic may affect multiple conversation turns, $L$ is the number of plans.

Given a designated target topic $z_{T^{'}}$ paired with its action $a_{T^{'}}$, a user profile $\mathcal{U}^{'}$, a set of relevant domain knowledge $\mathcal{K}^{'}$, and a conversation history $\mathcal{H}^{'}$, our objective is to generate coherent utterances to engage the user in the conversation so as to recommend $z_{T^{'}}$ when appropriate. Due to the complexity of the problem, it can be decomposed into three sub-tasks: (1) \textbf{action planning}, i.e., plan actions to determine where the conversation should go to lead the conversation proactively; (2) \textbf{topic planning}, i.e., plan appropriate topics to move forward to the target topic; (3) \textbf{dialogue generation}, i.e., generate a proper system utterance to achieve the planned action and topic at each turn.

\subsection{Our Method}
In this section, we introduce our TCP framework, which guides dialogue generation in a pipeline manner (see Figure \ref{fig:comp.c}). First, we use different encoders to learn representations of different types of inputs. Second, we propose a target-driven conversation planner to plan a path consisting of dialogue actions and topics. After planning, we adopt the planned content to guide dialogue generation.

\subsubsection{Encoders}

For the user profile $\mathcal{U}^{'}$, we adopt an end-to-end memory network \cite{sukhbaatar2015end} to encode $\mathcal{U}^{'}$, which is represented as $\mathbf{U}=(\mathbf{u}_{1},\mathbf{u}_{2},\cdots,\mathbf{u}_{m})$, $m$ is the length of the user profile. To efficiently represent the domain knowledge, we employ a Graph Attention Transformer \cite{galetzka-etal-2021-space} as the encoder, where knowledge triples are converted into unique relation-entity pairs instead of directly concatenating those triples in order to save space. Note that the embedding layers can be initialized from pre-trained language models (PLMs), e.g., BERT \cite{devlin-etal-2019-bert}. The final domain knowledge representation is denoted as $\mathbf{K}=(\mathbf{k}_{1},\mathbf{k}_{2},\cdots,\mathbf{k}_{k})$, where $k$ is the length of the domain knowledge. For the conversation history $\mathcal{H}^{'}$, we adopt a BERT \cite{devlin-etal-2019-bert} to encode $\mathcal{H}^{'}$, obtaining its token-level representation $\mathbf{H}=(\mathbf{h}_{1},\mathbf{h}_{2},\cdots,\mathbf{h}_{n})$, where $n$ is the length of $\mathcal{H}^{'}$.

\subsubsection{Target-driven Conversation Planner}

\begin{figure}[th!]
\centering
\includegraphics[width=0.96\linewidth]{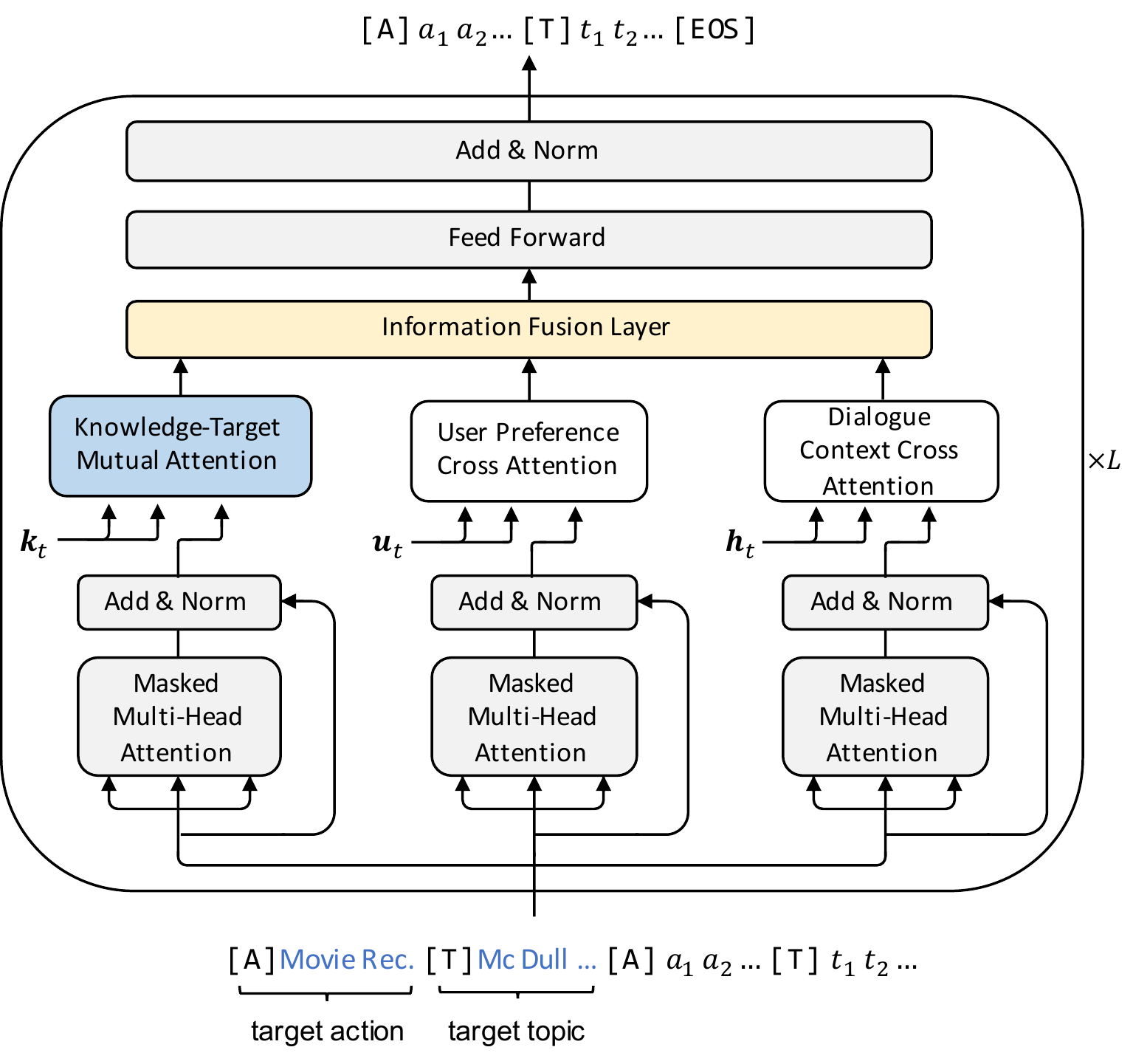}
\caption{Overview of Target-driven Conversation Planner.}
\label{fig:model}
\end{figure}

Our target-driven conversation planner aims to plan a path consisting of dialogue actions and topics in a generation-based manner. Since the target action and the target topic have been designated in advance and should also be bounded at the end of the path to be planned, we expect the target to drive the conversation planner to generate a more reasonable path. Intuitively, we let the conversation planner generate the path from the target turn of the conversation to the current turn (see Figure \ref{fig:comp.c}), which is of benefit to leverage more target-side information.
With such intuition, we build our target-driven conversation planner based on the Transformer \cite{vaswani2017attention} decoder architecture, which is shown in Figure \ref{fig:model}. It generates a plan sequence token by token, i.e., ``\texttt{[A]$a_{1}a_{2}\cdots$[T]$t_{1}t_{2}\cdots$[EOS]}''. Here, \texttt{[A]} is a special token to separate an action, \texttt{[T]} is a special token to separate a topic, \texttt{[EOS]} denotes the end of the plan sequence.

Concretely, to train the conversation planner, we put the tokens of the target action and the target topic ahead of the plan sequence as input (see Figure \ref{fig:model}), the hidden representation of which is denoted as $\mathbf{T}$.
During planning, the shifted token-level plan representation is used as the query. After being passed to three masked multi-head attention layers followed by add and normalization layers, it obtains the query representations $\mathbf{P}_{k}$, $\mathbf{P}_{u}$, and $\mathbf{P}_{h}$ with attentions to $\mathbf{K}$, $\mathbf{U}$, and $\mathbf{H}$, respectively.
Considering that the planned topics are mainly from the domain knowledge, and the target topic is essential to drive the entire conversation, we propose a \textbf{knowledge-target mutual attention} module. We use $\mathbf{K}$ and the target $\mathbf{T}$ to calculate a relevance score via the scaled dot-product \cite{vaswani2017attention}, the average of which can be viewed as a weight that the target influences the reasoning over the domain knowledge. When using $\mathbf{P}_{k}$ to attend to $\mathbf{K}$, the computation can be further given by:
\begin{align}
    \mathbf{K}_{weight}&=\text{MeanPooling}(\frac{\mathbf{K}\mathbf{T}^{\mathsf{T}}}{\sqrt{d}}) \\
    \mathbf{A}_{k}&=\text{softmax}(\frac{\mathbf{P}_k\mathbf{K}^\mathsf{T}}{\sqrt{d}}*\mathbf{K}_{weight})\mathbf{K}
\end{align}
where $\mathbf{A}_{k}$ is the attended representation, $d$ is the hidden size. At the same time, it is also important to consider the user preferences and the conversation progress during planning. Therefore, we use query representations $\mathbf{P}_{u}$ and $\mathbf{P}_{h}$ to attend to $\mathbf{U}$ and $\mathbf{H}$, obtaining $\mathbf{A}_{u}$ and $\mathbf{A}_{h}$, respectively. Both attentions are similar to the ``encoder-decoder cross attention'' in the Transformer \cite{vaswani2017attention} decoder. To leverage different parts of the attended results strategically, we add an information fusion layer through gate control, which is formulated as:
\begin{align}
\small
    \mathbf{A}_{1}&=\beta\cdot\mathbf{A}_{u}+(1-\beta)\cdot\mathbf{A}_{h} \\
    \beta&=\text{sigmoid}(\mathbf{W}_{1}[\mathbf{A}_u;\mathbf{A}_h]+\mathbf{b}_{1}) \\
    \mathbf{A}&=\gamma\cdot\mathbf{A}_{k}+(1-\gamma)\cdot\mathbf{A}_{1} \\
    \gamma&=\text{sigmoid}(\mathbf{W}_{2}[\mathbf{A}_k;\mathbf{A}_1]+\mathbf{b}_{2})
\end{align}
where $\mathbf{W}_{1},\mathbf{W}_{2}\in\mathbb{R}^{2d}$ are trainable parameters. $\mathbf{A}$ denotes the fused attended representation.

During training, we adopt the cross-entropy loss by comparing decoded plans and ground-truth plans. For inference, we employ the greedy search decoding to generate plan sequences.

\subsubsection{TCP-Enhanced Dialogue Generation}
\label{sec:pedg}

Since each planned path is in the order from the target turn to the current turn, we take the last action $a_{t}$ and the last topic $z_{t}$ in a path as the guiding prompt. Here, $z_{t}$ is further taken as the center topic to extract the corresponding triples from the domain knowledge, i.e., topic-centric attributes and reviews. They are expected to provide necessary knowledge beneficial to dialogue generation.
Note that we assume no domain knowledge is required when $a_{t}$ is ``chit-chat'', i.e., $z_{t}$ is ``\texttt{NULL}''.
Accordingly, we set the extracted knowledge as empty if this is the case. Finally, the concatenated text of the user profile, the extracted knowledge, the conversation history, and the action $a_{t}$ are taken as the input, and various backbone dialogue generation models can be applied to generate the system utterance. We will describe the backbone models we have adopted in Section \ref{baselines}.

\section{Experiments}

\subsection{Experimental Setup}

\subsubsection{Dataset}

We conduct experiments using the DuRecDial \cite{liu-etal-2020-towards-conversational} dataset, in which the system often proactively leads the conversation with rich interactive actions (e.g., chit-chat, question answering, recommendation, etc.). It contains about 10k multi-turn Chinese conversations and 156k utterances, with a sequence of dialogue actions and topics annotated for the system. Although there are several similar datasets such as \textsc{GoRecDial} \cite{kang-etal-2019-recommendation} and TG-ReDial \cite{zhou-etal-2020-towards}, we find that they are not very suitable for our experiments since their dialogues are mainly responding to users reactively.

We re-purpose the original DuRecDial dataset by automatic target creation. We regard the topic that the user has accepted at the end of each conversation as the target topic, and meanwhile the system's corresponding action is viewed as the target action (including movie/music/food/point-of-interest recommendations). In total, there are 15 actions and 678 topics (including a \texttt{NULL} topic). Following the splitting criterion in \citet{liu-etal-2020-towards-conversational}, we re-split the processed dataset into train/dev/test with 5,400/800/1,804 conversations, respectively. The number of turns is 7.9 on average, with a maximum of 14 turns. These conversations have an average of 4.5 different action/topic transitions from the start to the target.

\begin{table*}[th!]
\caption{Evaluation results of dialogue generation. The best result in terms of the corresponding metric is highlighted in boldface. Significant improvements over the backbone model results are marked with * (t-test, $p < 0.05$).}
\centering
\resizebox{0.96\textwidth}{!}{
\begin{tabular}{ll|r|c|c|c|c|c}
\toprule
\multicolumn{2}{c|}{\textbf{Model}} & \textbf{PPL ($\downarrow$)} & \textbf{F1 (\%)}  & \textbf{BLEU-1 / 2}  & \textbf{DIST-1 / 2}  & \textbf{Know. F1 (\%)} & \textbf{Target Succ. (\%)} \\
\midrule 
\multirow{4}{*}{Generation} & Transformer & 22.83 & 27.95 & 0.224 / 0.165 &  0.001 / 0.005 &  17.73 & 9.28 \\
  &  DialoGPT & 5.45 & 29.60 & 0.287 / 0.213 & 0.005 / 0.036 & 27.26 & 40.31 \\
  &  BART &  6.29 & 34.07 & 0.312 / 0.242  & \textbf{0.008} / 0.067  & 38.16  &  53.84 \\
  &  GPT-2 & 4.93 & 38.93 & 0.367 / 0.291 & 0.007 / 0.058 & 43.83 & 60.49 \\
\midrule
\multirow{2}{*}{Predict-then-generate} & MGCG\_G & 18.76 &  33.48 &  0.279 / 0.203  &  0.007 / 0.043 & 35.12 & 42.06 \\
  & KERS   & 12.55 & 34.04 & 0.302 / 0.220  & 0.005 / 0.030  & 40.75 &  49.40 \\
\midrule
\multirow{2}{*}{Ours}  & Ours (BART w/ TCP)  & 5.23 & 36.41* & 0.335* / 0.254* & \textbf{0.008} / \textbf{0.082} &  44.30* & 62.73* \\
  & Ours (GPT-2 w/ TCP) & \textbf{4.22} & \textbf{41.40}* & \textbf{0.376}* / \textbf{0.299}* & 0.007 / 0.072 &  \textbf{48.63}* & \textbf{68.57}* \\
\bottomrule
\end{tabular}}
\label{tab:dialogue}
\end{table*}

\subsubsection{Baseline Methods}
\label{baselines}

To validate our method, we first compare it with several competitive models for general dialogue generation: (1) \textbf{Transformer} \cite{vaswani2017attention}, which is a widely-used baseline model for language generation. (2) \textbf{DialoGPT} \cite{zhang-etal-2020-dialogpt}, which is a pre-trained dialogue generation model. (3) \textbf{BART} \cite{lewis2020bart}, which is an encoder-decoder pre-trained model for language generation. (4) \textbf{GPT-2} \cite{radford2019language}, which is a pre-trained autoregressive generation model. Note that we also employ the BART and GPT-2 as our backbone models for fine-tuning, following the description in Section \ref{sec:pedg} to conduct dialogue generation.
We also compare with state-of-the-art recommendation dialogue generation models, where they follow the predict-then-generate paradigm: (1) \textbf{MGCG\_G} \cite{liu-etal-2020-towards-conversational}, which employs the predicted next dialogue action and topic to guide the utterance generation. (2) \textbf{KERS} \cite{zhang-etal-2021-kers}, which has a knowledge-enhanced mechanism for recommendation dialogue generation.

To further explore the effect of planning for target-driven recommendation dialogue systems, we compare our TCP with (1) \textbf{MGCG} \cite{liu-etal-2020-towards-conversational}, which aims to perform multi-task predictions for the next dialogue action and topic. However, it assumes that ground-truth historical dialogue actions and topics are known for a system. In our problem formulation, we only provide the target action and topic, while the system itself should plan all interim dialogue actions and topics. We take the same input as our problem formulation for a fair comparison. (2) \textbf{KERS} \cite{zhang-etal-2021-kers}, which employs a Transformer \cite{vaswani2017attention} network to generate the next dialogue action and topic. Similarly, we take the same input as our problem formulation. (3) \textbf{BERT} \cite{devlin-etal-2019-bert}, which is fine-tuned by adding two fully-connected layers to jointly predict the next dialogue action and topic.

\subsubsection{Evaluation Metrics}

Following many previous studies, we adopt widely-used metrics including perplexity (\textbf{PPL}), word-level \textbf{F1}, \textbf{BLEU} \cite{papineni-etal-2002-bleu}, distinct (\textbf{DIST}) \cite{li-etal-2016-diversity}, and knowledge F1 (\textbf{Know. F1}) \cite{liu-etal-2020-towards-conversational}. In detail, the perplexity (\textbf{PPL}) and distinct (\textbf{DIST}) measure the fluency and the diversity of generated system utterances, respectively. The \textbf{F1} score estimates the precision and recall of generated utterances at the word level. The \textbf{BLEU} calculates $n$-gram overlaps between generated utterances and gold utterances.
The \textbf{Know. F1} evaluates the performance of generating correct knowledge (e.g., topics, attributes) from the domain knowledge triples. In particular, it is also essential to validate a model of how well the target topic is achieved. We choose the testing dialogues at the ``target turn'' to compute the ratio of generating the target topic correctly for each model, namely the target recommendation success rate (\textbf{Target Succ.}).
For conversation planning, following \citet{liu-etal-2020-towards-conversational}, we adopt accuracy (\textbf{Acc.}) to evaluate the predicted/generated action and topic for the next step. Due to the nature of conversations, multiple temporary planning strategies can be reasonable before completing the target. Following \citet{zhou2020augmenting}, we also expand ground-truth labels by taking the system's actions and topics within the previous turn and the following turn into account, formulating bigram accuracy (\textbf{Bi. Acc.}).

\subsubsection{Implementation Details}

Since the dataset is in Chinese, we adopt character-based tokenization. For TCP training, we use the pre-trained Chinese $\text{BERT}_{\text{base}}$ model, where the vocabulary size is 21,128 and the hidden size is 768. The target-driven conversation planner is stacked to 12 layers with 8 attention heads, using the same vocabulary with BERT, while the embeddings are randomly initialized. 
We adopt the Adam \cite{kingma2014adam} optimizer with an initial learning rate of $1e\text{-}5$. We train TCP for 10 epochs and warm up over the first 3,000 training steps with linear decay. We select the best model based on the performance on the validation set. For TCP inference, we adopt the greedy search decoding. For dialogue generation, we employ Chinese $\text{BART}_{\text{base}}$ and $\text{GPT-2}_{\text{base}}$ from the Huggingface's Transformers \cite{wolf-etal-2020-transformers} library as our backbone models. Each backbone model adopts the same parameter setting as that in baseline experiments. To boost the research in this direction, our code and data are publicly available \footnote{\url{https://github.com/iwangjian/Plan4RecDial}}.

\subsection{Results and Analysis}

\subsubsection{Evaluation Results}

Our evaluation results of dialogue generation are reported in Table \ref{tab:dialogue}. We observe that the vanilla Transformer performs inferior compared with other models since it has neither conversation planning nor pre-training. As pre-trained models, DialoGPT, BART, and GPT-2 can achieve much better performance over various metrics, which shows they are powerful to generate fluent and diverse utterances. For MGCG\_G and KERS, they achieve better results than Transformer and DialoGPT in terms of F1, BLEU, and knowledge F1. In view of the fact that MGCG\_G and KERS are trained without using pre-trained models, their improvements mainly benefit from the planning of the dialogue action and topic, which guides the system to generate more informative and more reasonable utterances. However, MGCG\_G and KERS obtain poor target recommendation success rates, which shows that they struggle to lead users towards the target topics when necessary. As shown in Table \ref{tab:dialogue}, with the benefit of our TCP, our models achieve significant improvements over all metrics, particularly with much higher target recommendation success rates. Evidently, our TCP-enhanced method is effective to guide the system to generate appropriate utterances.

\begin{table}[t!]
\caption{Experimental results of conversation planning. Significant improvements over the baseline models are marked with * (t-test, $p < 0.05$).}
\centering
\resizebox{0.9\linewidth}{!}{
\begin{tabular}{l|c|c|c|c}
\toprule
\multirow{2}{*}{\textbf{Model}} &  \multicolumn{2}{c|}{\textbf{Dialogue Action}} &  \multicolumn{2}{c}{\textbf{Dialogue Topic}} \\
 & \textbf{Acc. (\%)} &  \textbf{Bi. Acc. (\%)}  & \textbf{Acc. (\%)} & \textbf{Bi. Acc. (\%)} \\
\midrule
MGCG & 84.78 &  86.52 &  64.31  &  66.65 \\
KERS & 89.17  &  90.49  &  76.34   &  79.33  \\
BERT  & 90.19 &  91.35 &  83.53  & 85.61 \\
\midrule
TCP & \textbf{92.22}* & \textbf{93.82}* & \textbf{87.67}*  & \textbf{89.40}* \\
\bottomrule
\end{tabular}}
\label{tab:planning}
\end{table}

\subsubsection{Analysis of Conversation Planning}

To further validate the effect of planning for the formulated target-driven recommendation dialogue task, we compare TCP with other planning methods including MGCG, KERS, and BERT. The experimental results are reported in Table \ref{tab:planning}.
We observe that it is more difficult to predict/generate dialogue topics correctly than dialogue actions since the total size of the topics is much larger than that of the actions. Compared to the baseline methods, our TCP achieves substantial improvements in both dialogue action planning and topic planning. It verifies that TCP is able to plan an appropriate path consisting of proper dialogue actions and topics, which is effective to enable the system better understand what to say for the next step.

\section{Conclusion}
In this paper, we explore the target-driven recommendation dialogue task. We propose a Target-driven Conversation Planning (TCP) framework to proactively lead the conversation and guide dialogue generation. Experimental results demonstrate the effectiveness of our method. We will investigate how to plan more precisely and guide dialogue generation more effectively in the future.



\bibliographystyle{ACM-Reference-Format}
\bibliography{acmart}


\end{document}